%% file: main.tex
\documentclass{article}
\usepackage[preprint]{neurips_2024}

\author{
Nanbeige LLM Lab, Boss Zhipin
}

\usepackage{neurips_2024}

\usepackage{booktabs}
\usepackage{makecell}  

\usepackage[utf8]{inputenc} 
\usepackage[T1]{fontenc}    
\usepackage{hyperref}       
\usepackage{url}            
\usepackage{booktabs}       
\usepackage{amsfonts}       
\usepackage{nicefrac}       
\usepackage{microtype}      
\usepackage{xcolor}         
\usepackage{minted}

\usepackage{listings}
\usepackage{xcolor}  

\usepackage{tcolorbox}
\tcbuselibrary{listings, skins}
\tcbuselibrary{breakable}
\usepackage{algorithm}
\usepackage{algpseudocode}
\usepackage{amsmath}
\usepackage{url}
\usepackage{graphicx}
\usepackage{tikz}
\usetikzlibrary{arrows.meta,positioning,fit}
\usepackage{amsmath}
\usepackage{amsthm}
\usepackage{booktabs}
\usepackage{color}
\usepackage{xcolor}
\usepackage{xspace}
\usepackage[misc]{ifsym}

\usepackage[ruled, vlined, nofillcomment, linesnumbered, algo2e]{algorithm2e}

\usepackage{booktabs}
\usepackage{multirow}
\usepackage{balance}
\usepackage{enumitem}
\usepackage{stfloats}
\usepackage{diagbox}
\usepackage{fancyhdr}
\usepackage{stfloats}
\definecolor{result_color}{RGB}{250,250,210}
\usepackage{bm}

\newcommand{\ignore}[1]{}
\newcommand{\paratitle}[1]
{\vspace{1.5ex}\noindent\textbf{#1}}

\title{\large{Nanbeige4.2-3B: Unlocking Agentic Capabilities in a Compact Model}}

\begin{document}
\noindent\includegraphics[height=0.8cm]{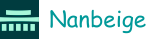}

\maketitle

\begin{abstract}
We present Nanbeige4.2-3B, a compact general agentic model with 3B non-embedding parameters. It delivers strong performance across code-agent, office-agent, and complex tool-use tasks while maintaining highly competitive reasoning capabilities in mathematics, coding, and science. Nanbeige4.2-3B is pretrained from scratch on 28T tokens with a Looped Transformer that reuses the layer stack to increase capacity without adding parameters. For SFT data and trajectory construction, we expand the diversity of executable environments, task assets, and agentic scaffolds through real-world deployment and large-scale synthesis. Our RL pipeline applies mixed-mode RLHF over Think and Non-Think responses to improve overall model quality and reduce failure cases, length-controlled reasoning RL to balance accuracy and reasoning efficiency, and agentic RL with outcome and process rewards to stabilize long-horizon training. Extensive evaluations show that Nanbeige4.2-3B outperforms larger models, including Qwen3.5-9B and Gemma4-12B, across diverse agentic benchmarks while remaining competitive on reasoning and alignment tasks. Performance with OpenClaw further supports its use as a compact local personal assistant. The model checkpoint is available at \url{https://huggingface.co/Nanbeige/Nanbeige4.2-3B}.
\end{abstract}

\begin{figure}[H]
 	\centering
 	\includegraphics[width=0.91\textwidth]{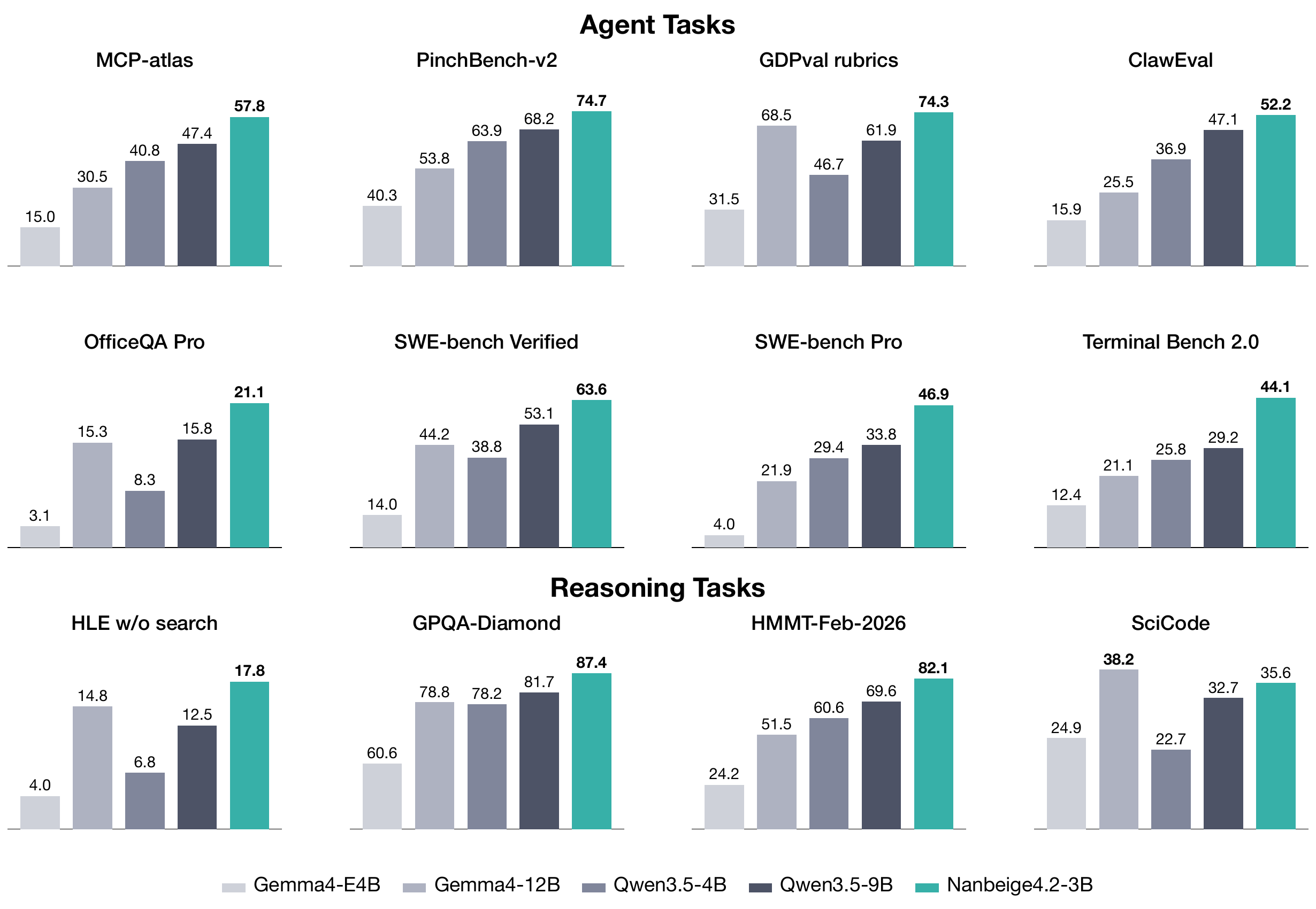}
	\caption{\textcolor{black}{Performance comparison between Nanbeige4.2-3B and other open-sourced models.
  }}
	\label{fig:thinking_performance}
 \end{figure}

\input{sec_introduction}
\input{sec_pretrain}
\input{sec_posttrain}

\input{sec_conclusion}

\newpage

\bibliographystyle{plain} 
\bibliography{references}
\newpage
\appendix
\input{sec_appendix}

\end{document}

%% file: sec_introduction.tex
\section{Introduction}\label{sec:intro}

Recent work shows that small language models can achieve strong performance on mathematical reasoning, competitive programming, and other verifiable tasks~\cite{xu2026vibethinker,yang2025nanbeige4}. Other studies investigate compact models that combine reasoning and tool use for specific agentic settings, such as deep search~\cite{yang2026nanbeige4}. However, strong agents must operate across heterogeneous environments, manipulate repositories and office documents, recover from failures, and make progress over long-horizon trajectories. Integrating these capabilities into one small model remains challenging.

Existing research typically equips small language models with advanced agentic capabilities through task-specific training. The resulting models often specialize in a single domain, such as repository-level coding or co-work tasks~\cite{ivison2026tmax,bai2026clawgym}, rather than serving as general-purpose agents. Although effective in their target environments, they leave open whether one compact model can support diverse agentic capabilities without sacrificing general reasoning ability. To investigate this question, we present \textbf{Nanbeige4.2-3B}, a compact model designed to maximize general-purpose agentic capabilities while retaining strong reasoning performance.

Rather than scaling parameters, 
Nanbeige4.2-3B adopts a Looped Transformer~\cite{bae2026mixture}. The architecture reuses the same Transformer stack to process hidden states for an additional pass, increasing effective model capacity without introducing additional parameters. 
Pretraining with this looped architecture from scratch on 28T tokens, with an expanded corpus and a refined data mixture, yields a strong base model. This provides a solid foundation for subsequent post-training stages.

After pretraining, we apply supervised fine-tuning (SFT) to develop broad agentic capabilities.
We first build a hybrid environment pool comprising real-world environments with a large collection of synthesized ones. 
Within these environments, we construct execution-grounded trajectories across repository-level software engineering, complex tool use, and artifact-centric office tasks, with diverse task assets and agentic scaffolds.
We then filter trajectories at both the trajectory and turn levels using execution signals and rubric-based assessments to maintain data quality at scale.

Building on the SFT model, we further develop a multi-stage RL pipeline that targets response quality, reasoning efficiency, and stable agentic learning. 
We first apply a two-stage RLHF across both Think and Non-Think responses to mitigate common failure modes such as hallucinations, repetitive generation, and instruction-following errors.
We then introduce length-controlled reasoning RL to improve reasoning efficiency by balancing concise reasoning and sufficient exploration. 
Finally, we perform agentic RL with combined outcome and process rewards to provide denser credit assignment and stabilize learning over complex trajectories.

The resulting model shows a broad agentic capability profile at its scale. Nanbeige4.2-3B achieves strong performance on complex tool-use, office-agent, and code-agent benchmarks, outperforming substantially larger Qwen3.5-9B~\cite{qwen3.5} and Gemma4-12B~\cite{team2026gemma} across diverse evaluations. It also remains competitive in mathematical, coding, and scientific reasoning, as well as alignment. When deployed within a general agent framework such as OpenClaw~\cite{openclaw}, it performs well on daily assistance, office workflows, and deep research, further supporting its use as a compact local assistant.

To the best of our knowledge, Nanbeige4.2-3B is the first open-source model at this scale to demonstrate this combination of code-agent, office-agent, and complex tool-use capabilities while retaining strong general reasoning performance. We open-source Nanbeige4.2-3B to support research on compact general agents and hope this release will encourage further study of how agentic capabilities can be developed under a strict parameter budget.

%% file: sec_pretrain.tex
\section{Pre-Training}

In this section, we first describe the Looped Transformer architecture, the choice of loop depth and initialization strategy, and the changes to our pre-training data recipe. Then we evaluate the pretrained base model against open-source models at a comparable parameter scale.

\subsection{Architecture}

To improve the effective depth and reasoning capacity under a fixed parameter budget, Nanbeige4.2-3B adopts a Looped Transformer architecture~\cite{bae2026mixture}. After the hidden states pass through the Transformer layers from bottom to top, they are fed through the same layer stack for an additional pass. It increases the effective computational depth and model capacity without introducing another set of parameters.

\paratitle{Training Looped Transformers from Scratch.}
One way to obtain a looped model is to first pretrain a standard transformer and then convert it into a looped architecture through upcycling~\cite{zhu2025scaling, yang2026iquest}. We compare this approach against training the looped architecture from scratch. In our experiments, the from-scratch approach performs significantly better, suggesting that the model benefits from adapting representations to repeated layer reuse throughout pre-training rather than introducing the loop only after a model has already been trained.

\paratitle{Loop Depth Selection.}
We study the number of passes through the shared layer stack. A two-pass configuration provides the most favorable trade-off in our experiments: relative to a standard Transformer, it retains approximately 75\% of the token efficiency and provides a significant capacity gain. Increasing the number of passes provides only marginal additional improvement, but substantially slows training and makes optimization less stable.

\paratitle{KV Cache Sharing and Scaling Configuration.}
To reduce the inference-memory overhead introduced by repeated computation, we investigate a variant that shares the KV cache across loop passes. Although the sharing configuration reduces the KV-cache by half, its performance gains are consistently lower than those of the full, non-sharing loop configuration. We therefore retain the full loop in Nanbeige4.2-3B to prioritize model performance. In addition to the loop configuration, we tune the depth of the transformer stack and its hidden width, seeking a balance between model capability and efficiency during both training and inference.  

\subsection{Data Recipe}

Our pre-training corpus comprises 28T tokens, exceeding that of Nanbeige 4.1 in both scale and data quality.
We further refine the data mixture by increasing the sampling weights of mathematics, code, and synthetic QA data, which we find particularly beneficial for compact models. We also incorporate a small proportion of agentic trajectory data into the pre-training mixture.
This mixture represents an initial step toward agentic pre-training. Further scaling the diversity, quality, and coverage of agentic data remains an important direction for future work.

\subsection{Pre-Training Evaluation}

We select a diverse set of benchmarks, including reasoning-oriented metrics (GSM8K~\cite{cobbe2021training}, BBH~\cite{suzgun2023challenging}, and MBPP~\cite{austin2021program}) and knowledge-oriented metrics (MMLU-Pro~\cite{wang2024mmlu}, SuperGPQA~\cite{du2026supergpqa}, and GPQA~\cite{rein2023gpqa}). Table~\ref{tab:pretrain_results} compares Nanbeige4.2-3B-Base with base models of a similar parameter scale, including Nanbeige4-3B-Base, Qwen3.5-4B-Base, and Gemma4-E4B-Base. Nanbeige4.2-3B-Base consistently improves upon Nanbeige4-3B-Base and performs significantly better than the other base models. The results demonstrate the effectiveness of combining the Looped Transformer structure with the improved pre-training data recipe.

\begin{table}[t]
    \centering
    \caption{Performance comparison of base models. 
    }
    \label{tab:pretrain_results}
    \small
    \setlength{\tabcolsep}{4pt}
    \renewcommand{\arraystretch}{1.05}
    \begin{tabular}{lccrrrrrr}
        \toprule
        & \multicolumn{2}{c}{\textbf{Model Size}}
        & \multicolumn{6}{c}{\textbf{Benchmark Scores}} \\
        \cmidrule(lr){2-3}
        \cmidrule(lr){4-9}
        \textbf{Model}
        & \textbf{Total}
        & \textbf{Non-emb.}
        & \textbf{GSM8K}
        & \textbf{BBH}
        & \textbf{MBPP}
        & \textbf{MMLU-Pro}
        & \textbf{SuperGPQA}
        & \textbf{GPQA} \\
        \midrule
        Gemma4-E4B
            & 8B & 4B
            & 61.8 & 62.5 & 53.5 & 37.6 & 23.3 & 27.5 \\
        Qwen3.5-4B
            & 5B & 4B
            & 84.4 & \underline{79.1} & 57.1
            & \underline{51.8} & \underline{32.1} & \underline{43.1} \\
        Nanbeige4-3B
            & 4B & 3B
            & \underline{85.9} & 70.7 & \underline{60.7}
            & 47.6 & 24.8 & 36.2 \\
        \midrule
        \textbf{Nanbeige4.2-3B}
            & \textbf{4B} & \textbf{3B}
            & \textbf{92.7} & \textbf{81.6} & \textbf{67.6}
            & \textbf{63.8} & \textbf{35.2} & \textbf{53.3} \\
        \bottomrule
    \end{tabular}
\end{table}

%% file: sec_posttrain.tex
\section{Post-Training}

In this section, we present the post-training pipeline used to develop the reasoning and agentic capabilities of Nanbeige4.2-3B. The pipeline combines scalable SFT data and trajectory construction with a multi-stage RL recipe designed to improve response quality, reasoning efficiency, and long-horizon agentic behavior. We then evaluate the resulting model across general and agentic benchmarks, as well as local personal-assistant scenarios.

\input{subsec_posttrain-data}

\input{subsec_training}

\input{subsec_experiment}

%% file: subsec_posttrain-data.tex
\subsection{Data}

We curate and synthesize a large-scale, multi-domain corpus of post-training data spanning Agentic Software Engineering, Complex Tool Use, and Agentic Cowork. By systematically expanding its task diversity and difficulty frontier, we aim to push compact models toward performance ceiling and probe the limits of capability that can be unlocked through high-quality, scalable supervision.

\subsubsection{Agentic Software Engineering: Repository-to-Trajectory Synthesis.}\label{agent-coding}

\begin{figure*}[t!]
    \centering
    \includegraphics[width=1\textwidth]{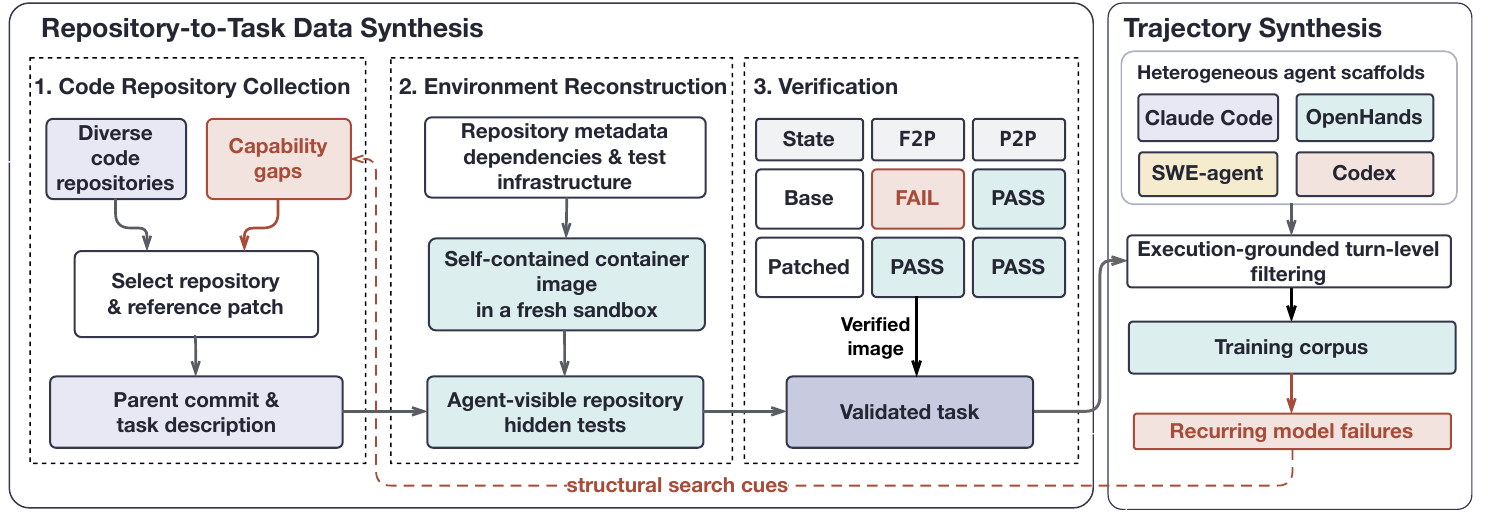}
    \caption{Agentic Software Engineering Data Synthesis Pipeline.}
    \label{fig:error2task_pipeline}
\end{figure*}

Unlike single-turn code generation, real-world agentic coding demands that models navigate complex repositories, infer intent from ambiguous specifications, localize defect paths, apply coordinated multi-file patches, and verify correctness via closed-loop execution. To construct high-quality trajectory supervision at scale, we establish a repository-level data synthesis pipeline. This system transforms historical GitHub development activities into executable tasks backed by automated environments, while leveraging model failure feedback to co-evolve task distribution alongside agent capabilities. The resulting repository-to-task data synthesis process is illustrated in Figure~\ref{fig:error2task_pipeline}.

\paratitle{Capability-Guided Code Repository Collection.}
To ensure the diversity and scale of our data, we first collect code repositories spanning programming languages, project scales, and application domains, from which we construct well-scoped software-engineering tasks with executable verification. 
To further identify repositories with the greatest potential to improve model capabilities, we run our model on a small set of seed training tasks, analyze and categorize its failure modes, and use the resulting capability gaps to guide repository and patch selection. 
This process preserves diversity across repositories and problem settings while improving coverage of capability areas in which the model exhibits systematic weaknesses.

\paratitle{Executable Environment Reconstruction and Task Synthesis.}
Once a repository is prepared and the patch is selected, we reconstruct the execution environment by analyzing repository metadata, dependency specifications, and test infrastructure. We then build a self-contained container image from the parent commit in an isolated sandbox.
Tasks are synthesized based on the selected patches, while the patches themselves and task-specific grading tests are withheld from the repository exposed to the agent.
Hidden tests are injected only during evaluation.

\paratitle{Verification and Closed-Loop Task Evolution.}
Before trajectory synthesis, we validate each task using an executable verifier constructed from the original source and test changes. 
Fail-to-pass tests confirm that the target behavior is absent in the base repository and restored by the reference patch, while pass-to-pass tests protect existing functionality from regression. 
We also audit whether the task description, hidden tests, and reference patch express a consistent behavioral contract, revising recoverable tasks and discarding broken, flaky, or underspecified ones. 
During routine inference and training, recurring model failures are automatically collected and converted into structural search cues for subsequent repository mining. 
This closed-loop mechanism expands the task space to systematically concentrate on weaknesses detected during the model’s own evolution.

\paratitle{Trajectory Synthesis across Diverse Scaffolds.}
To prevent model over-fitting to specific prompt templates or tool-use conventions, we deploy multiple heterogeneous agent scaffolds within our containerized sandbox, including Claude Code, OpenHands, SWE-agent, and specialized Codex-based drivers.
Given the same repository state and fail-to-pass verification tasks, each agent independently explores and resolves problems in parallel. 
Their distinct system prompts, context compression policies, and editing interfaces induce diverse solution strategies (e.g., broad exploratory search vs. localized pinpoint editing). 
Aggregating these complementary trajectories yields a training corpus that emphasizes scaffold-invariant reasoning and repair strategies, rather than scaffold-specific interaction patterns, thereby improving generalization to unseen agent interfaces and tool schemas.

\paragraph{Execution-Grounded Turn-Level Filtering.}
To ensure data quality, we retain only trajectories whose resulting patches pass the designated target and regression tests. 
We further conduct a turn-level filtering for incorrect tool calls, non-terminating loops, redundant actions, and context truncations, prioritizing concise trajectories while preserving essential debugging and recovery steps. 
Finally, all trajectories are normalized into a unified multi-turn schema comprising observed assistant messages, tool calls, and environment feedback.

\subsubsection{Tool Use: Scaling Trajectory Synthesis with Hybrid Real-Simulated Environments.} 

\begin{figure*}[t!]
	\centering
	\includegraphics[width=1\textwidth]{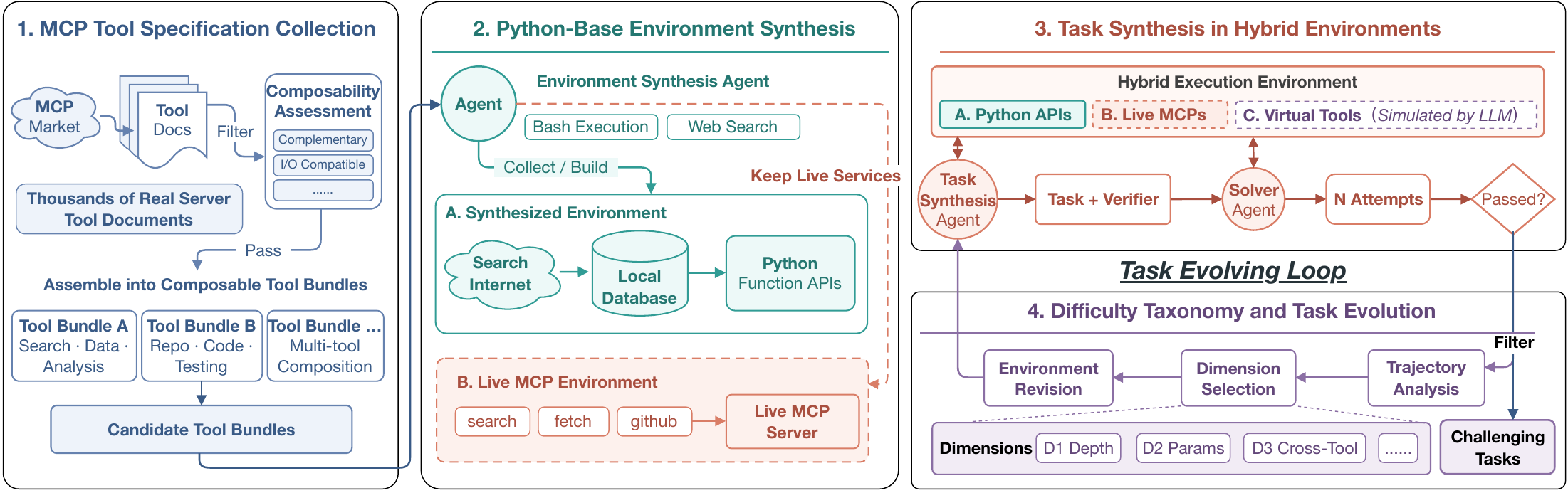}
	\caption{\textcolor{black}{Tool Use Data Synthesis Pipeline.
 }}
	\label{fig:mcp}
\end{figure*}

We collect and synthesize a large corpus of high-quality tool-use trajectories.
Building upon the data-synthesis pipeline developed for Nanbeige4.1~\cite{yang2025toolmind}, we further scale the diversity and complexity of tool-use tasks by constructing hybrid environments that integrate real-world APIs, executable tool interfaces, and model-simulated environmental components.
The resulting framework comprises four components as illustrated in Figure~\ref{fig:mcp}: tool specification collection, executable environment synthesis, task synthesis, and a difficulty taxonomy that supports automated labeling and adaptive evolution.

\paratitle{MCP Tool Specification Collection.}
We collect thousands of tool specification documents for real-world MCP servers from MCP marketplaces. For each, we assess its composability---that is, whether it can be orchestrated with other tools to support complex, multi-step tool-use scenarios. 
Then, tools with sufficient compositional potential are grouped into bundles that serve as the foundation for downstream task construction.

\paratitle{Python-Based Executable Environment Synthesis.}
For each tool-documentation bundle, an environment-synthesis agent equipped with web-search and Bash-execution capabilities automatically acquires relevant real-world data from the Internet and persists it in a local database. It then uses Python to reconstruct each tool as a callable function interface, thereby creating a self-contained execution environment that combines real data with a complete tool suite.

\paratitle{Task Synthesis in Hybrid Environments.}
Beyond Python-based executable environments, we retain live MCP services for tools such as web search, web scraping, and code-repository access, whose dynamic and time-sensitive outputs cannot be faithfully captured by static databases. For tools that depend on extensive runtime environments and cannot be readily reproduced in Python, we instead use a model to simulate their behavior and outputs.
Accordingly, our synthesis pipeline integrates three complementary environment types: (1) live online MCP services, (2) local tools implemented in Python, and (3) model-simulated virtual tools. This design supports both customized tool-use scenarios with complex data dependencies and general-purpose tool use requiring real network interaction, balancing environmental diversity with ecological realism.

\paratitle{Difficulty Taxonomy and Adaptive Task Evolution.}
To ensure that synthesized tasks attain sufficient difficulty and complexity, we predefine a difficulty taxonomy that systematically characterizes tasks along multiple dimensions, such as tool-use chain depth, information-retrieval difficulty, and parameter-inference complexity. Conditioned on the current environment, a task-synthesis agent with access to Bash execution generates natural-language task specifications together with Python verification functions for programmatic correctness evaluation.
After synthesis, a solver agent attempts each task multiple times. Trajectories from reliably solved tasks are fed back to the task-synthesis agent to identify the model's capability frontier and selectively evolve tasks by increasing difficulty along the relevant dimensions, thereby establishing an iterative closed-loop process.

\paratitle{Rapid Validation Construction.}
\label{sec:rapid_validation}
In parallel, we construct a compact held-out task suite for rapid validation during development. The suite covers representative tool bundles, hybrid environment types, and difficulty levels, while remaining disjoint from the training data. It enables fast and reproducible assessment of whether successive data-synthesis and reinforcement-learning iterations improve tool-use performance before full-scale evaluation.

\subsubsection{Agentic Cowork: Scaling Complex Office Tasks through Artifact-Centric Curation.}

\begin{figure*}[t!]
	\centering
	\includegraphics[width=1\textwidth]{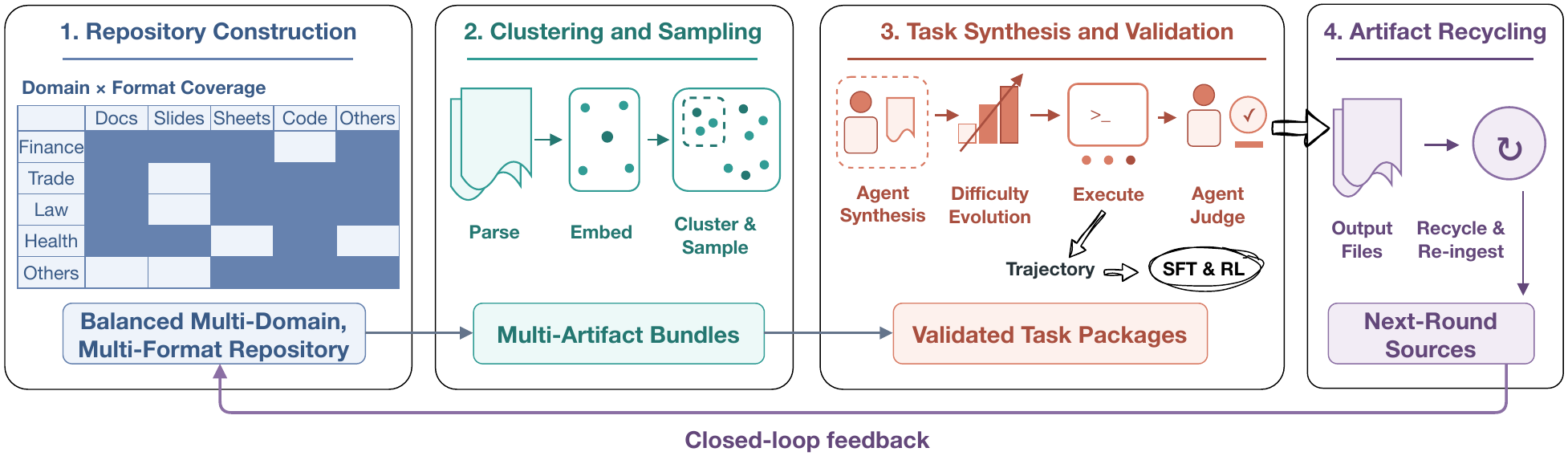}
	\caption{\textcolor{black}{Agentic Cowork Data Synthesis Pipeline.
 }}
	\label{fig:agentic_cowork_pipeline}
\end{figure*}

Beyond foundational tool use, agents must handle specialized office workflows involving reports, slide decks, spreadsheets, and other artifacts—capabilities central to knowledge-worker productivity. 
To equip foundation models with this agentic cowork capabilities, we develop an artifact-centric curation pipeline that synthesizes, evolves, and recycles task artifacts to scale diverse and complex real-world office workflows. 
In detail, the pipeline comprises artifact repository construction, vector-similarity-based domain clustering, task synthesis, trajectory synthesis, and closed-loop artifact reuse, as illustrated in Figure~\ref{fig:agentic_cowork_pipeline}.

\paratitle{Artifact Repository Construction.}
We construct a balanced artifact repository by collecting materials from diverse professional domains, including finance, trade, law, healthcare, and others. The repository covers heterogeneous formats, such as reports, slide decks, spreadsheets, PDFs, emails, and others, while calibrating their relative proportions to achieve broad and realistic coverage.

\paratitle{Embedding-Based Domain Clustering.}
We parse and normalize the collected artifacts into structured representations, encode them into a shared embedding space, and cluster them according to semantic and domain similarity. We further group and sample artifacts within each cluster to construct coherent yet diverse multi-artifact bundles for downstream task synthesis.

\paratitle{Task, Rubric, and Trajectory Synthesis.}
A task-synthesis agent interactively explores and manipulates selected artifact bundles within a sandboxed workspace to synthesize office tasks and their evaluation rubrics. We assign difficulty tiers and iteratively evolve tasks along multiple dimensions, including cross-artifact dependencies, tool-use requirements, workflow length, constraints, and verification demands. Open-source models subsequently execute the resulting tasks to generate tool-use trajectories and deliverable artifacts. An independent judge agent then evaluates the fidelity of each deliverable to the task specification, the consistency of the associated rubric, and the overall quality of the execution trace, retaining only high-quality samples for subsequent synthesis rounds.

\paratitle{Closed-Loop Artifact Recycling.}
Validated deliverables are re-ingested into the artifact repository as source materials for subsequent synthesis rounds. This closed-loop process progressively expands the task distribution, enabling the generation of increasingly diverse and complex office workflows.

%% file: subsec_training.tex
\subsection{Training Recipe}
Our training recipe consists of four stages: supervised fine-tuning~(SFT), two-stage RLHF for hybrid thinking, reasoning RL with length control, and agentic RL with action-centric rubrics.

\subsubsection{SFT}
\begin{figure*}[t!]
    \centering
    \includegraphics[width=0.8\textwidth]{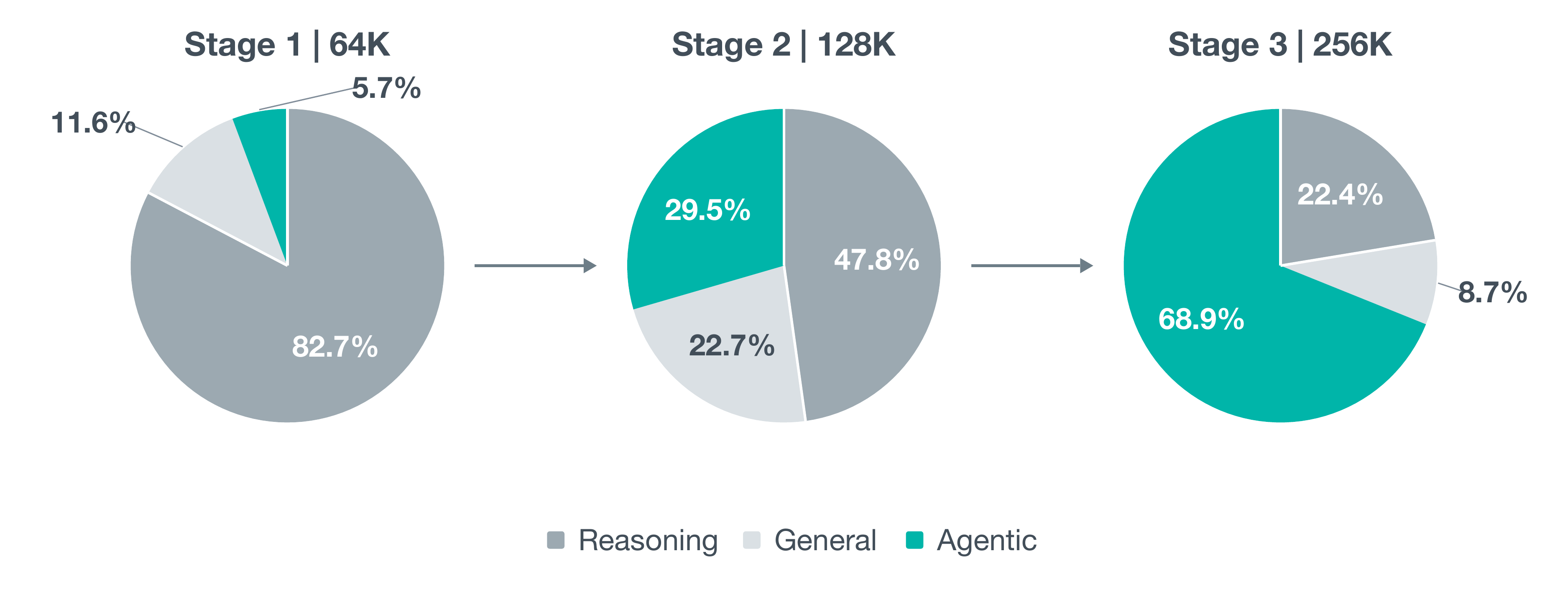}
    \caption{Stage-wise SFT target-token mixtures for the 64K, 128K, and 256K stages.}
    \label{fig:sft-stage-mixture}
\end{figure*}

Starting from the pretrained checkpoint, we conduct supervised fine-tuning with a three-stage curriculum that progressively extends the maximum training context from 64K to 128K and 256K tokens. Rather than maintaining a fixed data mixture, we gradually shift the distribution of supervised target tokens from STEM-centered reasoning toward long-horizon agentic interaction as the model develops a stronger reasoning foundation.

\paratitle{Three-Stage Curriculum.}
In the 64K stage, the supervised target-token mixture is dominated by reasoning-oriented STEM data (82.7\%), with 11.6\% general instruction data and 5.7\% agentic data. This stage establishes the foundation for scientific, mathematical, and engineering problem solving. In the 128K stage, the mixture shifts to 47.8\% reasoning-oriented STEM data, 22.7\% general instruction data, and 29.5\% agentic data, providing a transition toward longer-context instruction following, tool use, and multi-turn environment interaction.
After the model acquires a sufficiently strong reasoning foundation, the 256K stage shifts the primary training objective toward agentic capability. We allocate 68.9\% of the supervised target tokens to agentic data, while retaining 22.4\% reasoning-oriented STEM data and 8.7\% general instruction data. This strengthens long-horizon planning, tool invocation, execution feedback utilization, and recovery from intermediate failures. The three-stage mixtures are summarized in Figure~\ref{fig:sft-stage-mixture}.

\paratitle{Turn-Level Loss Masking.}
A successful trajectory may still contain incorrect or unproductive intermediate turns. We assign a binary loss mask $m_t$ to each assistant turn using execution feedback and rubric-based assessment. Unreliable turns are excluded from the SFT loss ($m_t=0$) but retained, together with their resulting observations, in the context of subsequent turns. This preserves the full interaction context without training on incorrect turns, allowing the model to learn how to recover from mistakes in long-horizon tasks.

\subsubsection{Two-Stage RLHF for Hybrid Thinking}

After mixed-domain SFT, the compact model still exhibits unstable generation behaviors, including repetitive reasoning, cyclic reflection, delayed termination, duplicated answers, malformed formats, and abnormal continuations. To address these failure cases and establish a stable foundation for subsequent reasoning and agentic RL, we first apply a two-stage RLHF procedure covering both think and non-think responses. 

We train a point-wise reward model that independently evaluates the overall
quality of each response. 
The reward favors responses that are correct, relevant, well formatted, and
properly terminated, while discouraging repetition, ineffective continuation,
and malformed outputs. In addition, the reward takes both safety and user-friendliness into account to ensure that the model maintains harmless and friendly interactions while adhering to the specified generation behavior.
A central finding is that the behavioral regularization learned
from general-purpose RLHF exhibits strong generalization along two distinct
dimensions:

\textbf{Cross-task generalization.}
General-purpose RLHF improves not only instruction following but also
mathematical reasoning, coding, and agentic performance. This observation
challenges the conventional view of RLHF as primarily a mechanism for
controlling safety and conversational style. Instead, behaviors such as
avoiding repetition, respecting output formats, and terminating appropriately
act as a form of trajectory-level regularization. In complex agent tasks, many
failures are caused not by incorrect reasoning, but by generation loops,
malformed outputs, or abnormal continuations that interrupt downstream
execution. Removing these failures allows the model's existing reasoning
capabilities to be expressed more reliably.

\textbf{Cross-mode generalization.}
Behavioral constraints learned predominantly from Non-Think responses generalize
directly to Think-mode generation. Although the second stage mainly trains on non-think responses to general queries, it also markedly
reduces cyclic reflection, redundant reasoning, and abnormal continuation in
Think mode. This indicates that structured generation, repetition control, and
termination awareness are not mode-specific skills, but shared capabilities
underlying both response modes. 

To provide a concise illustration of these improvements,
Table~\ref{tab:rlhf-main-results} reports representative results across three
capability categories: AA-LCR\footnote{https://huggingface.co/datasets/ArtificialAnalysis/AA-LCR} for alignment, LiveCodeBench-V6~\cite{jain2024livecodebenchholisticcontaminationfree} for reasoning,
and PinchBench-V2\footnote{https://pinchbench.com/} for agentic capability. For each benchmark, we compare accuracy,
average output length, and bad-case rate (percentage of data with formatting errors or exceeding the length limit) across the SFT, Think-RLHF, and final
Think-to-Non-Think RLHF checkpoints. All checkpoints are evaluated in Think
mode under the same decoding configuration. These three benchmarks
 consistently show that the two-stage RLHF procedure improves task performance while producing
shorter and more stable generation trajectories.

\begin{table*}[t]
    \centering
    \small
    \setlength{\tabcolsep}{4pt}
    \renewcommand{\arraystretch}{1.10}

    \begin{tabular}{@{}llrrr@{}}
        \toprule
        \multirow[c]{2}{*}{\textbf{Capability}}  & \multirow[c]{2}{*}{\textbf{Metric}}   & \multicolumn{3}{c}{\textbf{Training Stage}} \\
        \cmidrule(lr){3-5}  
        & & \textbf{SFT}  & \textbf{Think RLHF}  & \textbf{Non-Think RLHF} \\
        \midrule

        \multirow{3}{*}{\makecell[l]{Alignment\\\textit{AA-LCR}}}
        & Accuracy (\%) $\uparrow$
        & 50.00 & 53.00 & \bfseries 57.00 \\
        & Bad-case Rate (\%) $\downarrow$
        & 17.00 & 8.00 & \bfseries 2.00 \\
        & Average Length (tokens) $\downarrow$
        & 19,545 & 13,791 & \bfseries 7,915 \\
        \midrule
        
        \multirow{3}{*}{\makecell[l]{Reasoning\\\textit{LiveCodeBench-V6}}}
        & Accuracy (\%) $\uparrow$
        & 65.45 & 68.51 & \bfseries 72.10 \\
        & Bad-case Rate (\%) $\downarrow$
        & 6.49 & 0.95 & \bfseries 0.00 \\
        & Average Length (tokens) $\downarrow$
        & 25,905 & 16,781 & \bfseries 15,182 \\
        \midrule
        
        \multirow{3}{*}{\makecell[l]{Agentic\\\textit{PinchBench-V2}}}
        & Accuracy (\%) $\uparrow$
        & 55.89 & 71.14 & \bfseries 75.49 \\
        & Bad-case Rate (\%) $\downarrow$
        & 5.44 & 8.16 & \bfseries 2.72 \\
        & Average Length (tokens) $\downarrow$
        & 20,515 & 13,098 & \bfseries 11,808 \\
        
        \bottomrule
    \end{tabular}

    \caption{
    Representative effects of two-stage RLHF on alignment, reasoning, and
    agentic capabilities. All checkpoints are evaluated in Think mode under
    identical decoding settings.
    }
    \label{tab:rlhf-main-results}
\end{table*}

\subsubsection{Reasoning RL with Length Control}

 Following general-purpose RLHF, we further conduct reasoning RL to strengthen Think-mode capability. Unlike Nanbeige4.1, where reasoning RL optimized primarily for task performance, this version pursues accuracy gains while simultaneously constraining reasoning length. To this end, we introduce a problem-dependent length-control objective for Think-mode RL. The objective combines a fixed length budget with a continuous, difficulty-aware penalty, encouraging concise reasoning on reliably solved problems while preserving exploration on difficult ones.

\textbf{Offline budget construction.}
For each problem $q$, we collect successful historical rollouts from checkpoints preceding this stage and set its length budget $b_q$ to the median length of these correct responses. The budgets are computed once and remain fixed throughout the training process.

\textbf{Difficulty-aware length penalty.}
Training alternates between a \emph{constrained} phase, which penalizes
over-long responses, and a \emph{free-expansion} phase, which disables the
penalty to preserve exploration. The steps in each phase are two
separate hyperparameters. In constrained phase, given a response of
length $L_i$, maximum rollout length $L_{\max}$, and base task reward
$r_i^{\mathrm{base}}$, we shape the final reward as:
\begin{equation}
r_i = r_i^{\mathrm{base}} - \alpha\, p_q
\left[\frac{L_i-b_q}{L_{\max}-b_q}\right]_0^1,
\end{equation}
where $[x]_0^1=\min(\max(x,0),1)$, $\alpha>0$ bounds the penalty magnitude, and
$p_q$ is the fraction of fully correct responses in the current rollout group.
The penalty is thus difficulty-aware: responses within budget are never
penalized, while the penalty for over-budget responses grows with both the normalized excess length and the problem's current pass rate, so concise reasoning is encouraged on reliably solved problems but not on still-difficult ones.
Because the penalty is subtracted rather than gating the task reward, it
preserves correctness differences and remains continuous at the budget
boundary; for binary rewards, $\alpha<1$ further guarantees that a correct
response is always preferred to an incorrect one regardless of length.

\subsubsection{Agentic RL with Action-Centric Rubrics.}

Across diverse agentic settings, including general-purpose tool use, search, cowork, and coding, we design action-evaluation rubrics to diagnose and assess behavioral deficiencies in agents. These rubrics measure dimensions such as tool-call accuracy and the information gain each turn contributes toward task completion, and their scores serve as process-level rewards that penalize recurrent action errors at the turn level.

Data selection is equally important at this scale. Using the reasoning-RL model to estimate task pass rates, we restrict agentic RL to relatively easy tasks, namely those with short trajectories and relative higher pass@8 score. For a compact model, we find that such tasks yield both more stable optimization and larger performance gains than harder or longer-horizon ones.

\begin{figure}
    \centering
    \includegraphics[width=1\linewidth]{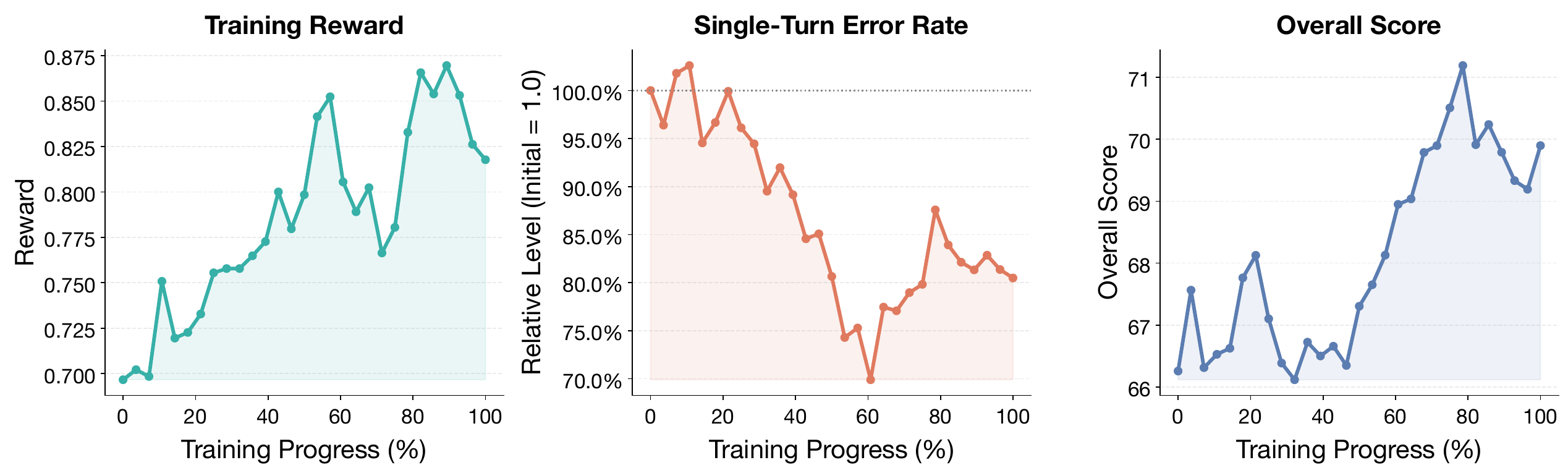}
    \caption{Training Dynamics with Action-Centric Agentic RL.}
    \label{fig:training_dynamics}
\end{figure}

With this data and reward design, we run agentic RL and track its dynamics on the Rapid Validation Tasks mentioned in Sec~\ref{sec:rapid_validation}. We report the single-turn action error rate, normalized by the initial checkpoint, together with the overall task score, as shown in Figure~\ref{fig:training_dynamics}. Despite fluctuations caused by the diversity of trajectories and task difficulty, the training reward exhibits an overall upward trend. Meanwhile, the normalized action error rate decreases by approximately 20\%, while the overall score improves from 66.0 to a peak of 71.0 during training. These results show that combining outcome and action-centric process rewards suppresses recurrent action errors and improves end-task performance.

\subsubsection{Performance Gains from Reinforcement Learning}

\begin{figure}
    \centering
    \includegraphics[width=0.9\linewidth]{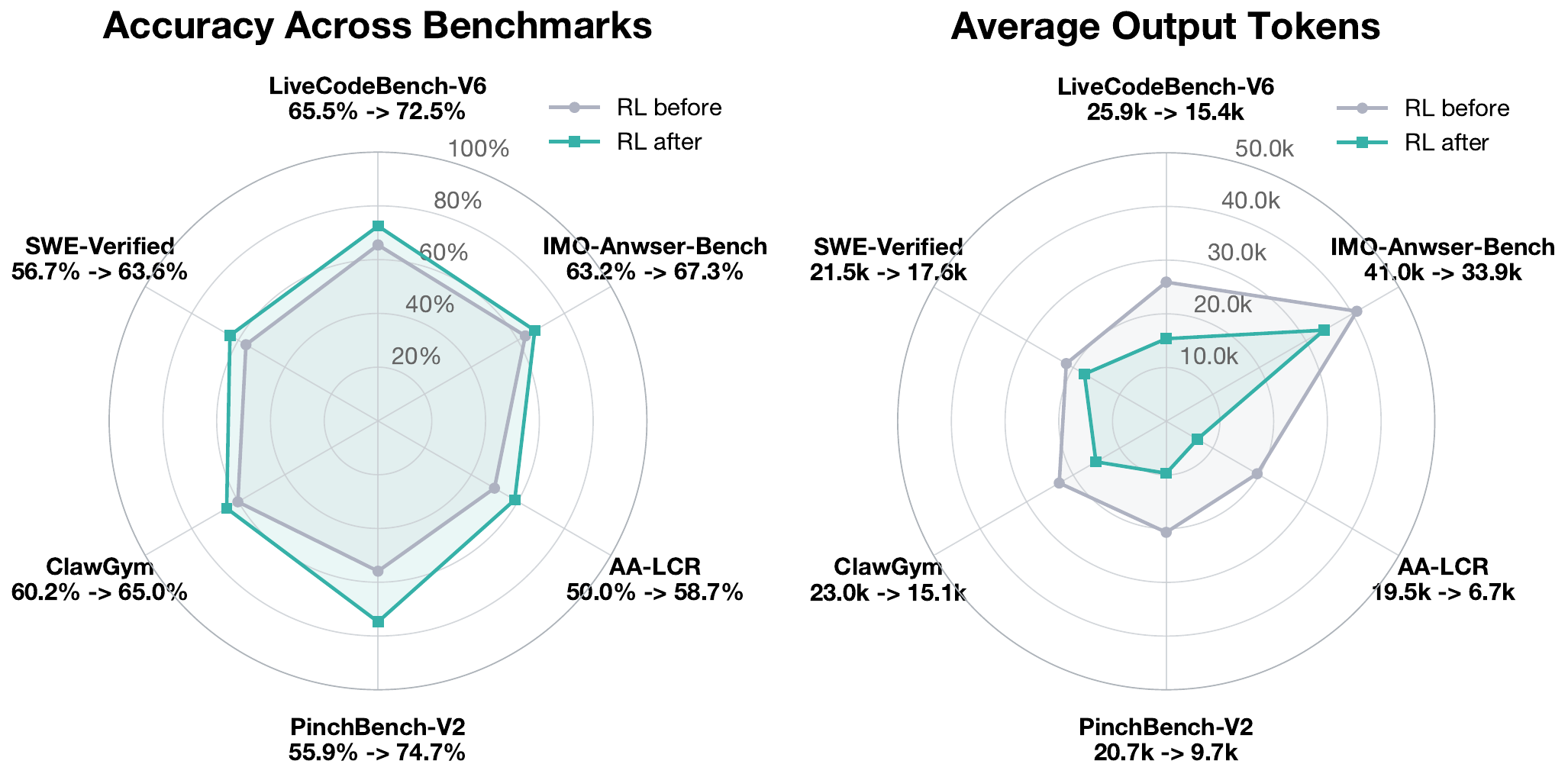}
    \caption{Accuracy and average token usage before and after RL across diverse benchmarks.}
    \label{fig:rl_acc_token_radar}
\end{figure}

To characterize the cumulative effect of the complete RL pipeline, we evaluate
the pre-RL and post-RL checkpoints on six benchmarks covering
mathematical and code reasoning, alignment, software engineering, and agentic
task execution.
As illustrated in Figure~\ref{fig:rl_acc_token_radar}, among these benchmarks, the post-RL checkpoint achieves higher accuracy while using fewer output tokens.

%% file: subsec_experiment.tex
\subsection{Evaluation}

We evaluate the model from two complementary perspectives: \textbf{(1) general and agentic capabilities} and \textbf{(2) local personal assistant performance}.
The first component assesses the model across complex tool-use, coding, and office-agent tasks, as well as mathematical and scientific reasoning, competitive programming, and instruction following. Together, these benchmarks characterize the model's agentic, reasoning, and alignment capabilities under standardized evaluation protocols.
The second component focuses on practical local assistant applications. Rather than measuring individual capabilities in isolation, these benchmarks evaluate the model within a unified agent framework on end-to-end tasks involving multi-step planning, tool interaction, office productivity, and deep research. This setting more closely reflects how the model would be used in local deployments.
Together, the two components provide a comprehensive view of the model’s general capability profile and its effectiveness in practical agentic applications.

\subsubsection{General and Agentic Capabilities}

We evaluate Nanbeige4.2-3B across a diverse benchmark suite with four major capability categories:

\begin{itemize}[leftmargin=10pt]

\item \textbf{General Agents.}
GDPval Rubrics~\cite{fry2025gdpval}, AgentIF-Oneday~\cite{chen2026agentifoneday}, OfficeQA-Pro~\cite{opsahlong2026officeqa}, PinchBench-V2, ClawGym~\cite{bai2026clawgym}, Claw-Eval~\cite{ye2026claw}, and MCP-Atlas~\cite{bandi2026mcpatlas} assess general agent capabilities, including office productivity, multi-step task execution, tool use, and realistic workflow completion.

\item \textbf{Code Agents.}
SWE-Bench Verified~\cite{jimenez2024swebench}, SWE-Bench Pro~\cite{deng2025swebenchpro}, and Terminal-Bench 2.0~\cite{merrill2026terminalbench} evaluate software engineering performance through repository-level code modification, debugging, terminal interaction, and autonomous development tasks.

\item \textbf{Reasoning.}
HLE~\cite{phan2025humanitysexam}, SciCode~\cite{tian2024scicode}, GPQA-Diamond, HMMT-Feb-2026~\cite{dekoninck2026matharena}, IMO-Answer-Bench~\cite{luong2025towards}, and LiveCodeBench-V6 cover scientific, mathematical, and code reasoning.

\item \textbf{Alignment.}
AA-LCR and IF-Bench~\cite{pyatkin2025generalizing} assess long-context reasoning and complex instruction-following abilities. We also include Recruit-Bench, our in-house benchmark that covers both enterprise hiring (B2C) and candidate job-seeking (C2B) scenarios. It evaluates model alignment in realistic enterprise recruitment workflows. 

\end{itemize}

For comparison, we include open-source models of similar or larger sizes, including Qwen3.5-4B, Qwen3.5-9B, Gemma4-E4B, and Gemma4-12B. Detailed evaluation settings and benchmark-specific protocols are provided in the Appendix~\ref{app:evaluation}.

\begin{table*}[t]
\centering
\small
\setlength{\tabcolsep}{5.0pt}
\renewcommand{\arraystretch}{1.08}
\caption{Evaluation results across general-agent, code-agent, reasoning, and alignment benchmarks.}

\label{tab:general_agentic}
\begin{tabular}{lccccc}
\toprule

& \makecell[c]{Qwen3.5-4B}
& \makecell[c]{Qwen3.5-9B}
& \makecell[c]{Gemma4-E4B}
& \makecell[c]{Gemma4-12B}
& \makecell[c]{Nanbeige4.2-3B} \\
\midrule
\# Total Params        & 5B    & 10B   & 8B    & 12B   & 4B \\
\# Non-embedding Params& 4B    & 8B    & 4B    & 10B   & 3B \\
\midrule

\multicolumn{6}{c}{\textit{General Agent}}\\
\midrule

GDPval Rubrics     & 46.7 & 61.9 & 31.5 & \underline{68.5} & \textbf{74.3}\\
AgentIF-Oneday    & 56.9 & \underline{60.4} & -- & -- & \textbf{67.5}\\
OfficeQA-Pro      & 8.3 & \underline{15.8} & 3.1 & 15.3 & \textbf{21.1}\\
PinchBench-V2     & 63.9 & \underline{68.2} & 33.3 & 53.8 & \textbf{74.7}\\
ClawGym           & 53.0 & \underline{56.1} & 16.4 & 40.8 & \textbf{65.0}\\
Claw-Eval $_{\text{\tiny Pass\^{}3}}$        & 36.9 & \underline{47.1} & 15.9 & 25.5 & \textbf{52.2}\\
MCP-Atlas          & 40.8 & \underline{47.4} & 15.0 & 30.5 & \textbf{57.8}\\

\midrule

\multicolumn{6}{c}{\textit{Code Agent}}\\
\midrule

SWE-Bench Verified & 38.8 & \underline{53.1} & 14.0 & 44.2 & \textbf{63.6}\\
SWE-Bench Pro      & 29.4 & \underline{33.8} & 4.0 & 21.9 & \textbf{46.9}\\
Terminal-Bench 2.0 & 25.8 & \underline{29.2} & 12.4 & 21.1 & \textbf{44.1}\\

\midrule

\multicolumn{6}{c}{\textit{Reasoning}}\\
\midrule

HLE (w/o Search)    & 6.8 & 12.5 & 4.0 & \underline{14.8} & \textbf{17.8}\\
SciCode            & 22.7 & {32.7} & 24.9 & \textbf{38.2} & \underline{35.6}\\
GPQA Diamond       & 78.2 & \underline{81.7} & 60.6 & 78.8 & \textbf{87.4}\\
HMMT-Feb-2026      & 60.6 & \underline{69.6} & 24.2 & 51.5 & \textbf{82.8}\\
IMO-Answer-Bench   & 46.8 & \underline{56.3} & 24.0 & 54.5 & \textbf{67.3}\\
LiveCodeBench-V6   & 55.8 & {65.6} & 55.3 & \underline{72.0} & \textbf{72.5}\\

\midrule

\multicolumn{6}{c}{\textit{Alignment}}\\
\midrule

AA-LCR             & 52.0 & \underline{58.0} & 30.7 & 55.3 & \textbf{58.7}\\
IF-Bench           & 41.4 & 54.1 & 44.0 & \textbf{73.5} & \underline{54.6}\\
Recruit-Bench      & 40.7 & 59.0 & 57.9 & \textbf{69.4} & \underline{63.3}\\

\bottomrule
\end{tabular}
\label{tab:overall_benchmarks}
\end{table*}

\paragraph{Overall Results.}

Table~\ref{tab:overall_benchmarks} summarizes the overall evaluation results. With only 3B non-embedding parameters, Nanbeige4.2-3B achieves the best results on all reported general-agent and code-agent benchmarks, outperforming the larger Qwen3.5-9B and Gemma4-12B models. These results cover complex tool-use, office workflows, repository-level software engineering, and terminal-based tasks, demonstrating broad agentic capability rather than specialization in a single environment.

Nanbeige4.2-3B also achieves the best result on five of the six reasoning benchmarks while remaining competitive on alignment tasks. Gemma4-12B performs better on IF-Bench and Recruit-Bench, but Nanbeige4.2-3B provides a stronger overall balance across agentic, reasoning, and alignment evaluations under a substantially smaller parameter budget.

\subsubsection{Local Personal Assistant}

With the small parameter scale, Nanbeige4.2-3B is compact enough for efficient local deployment while retaining strong reasoning and tool-use capabilities. To evaluate its effectiveness as a practical local assistant, we choose OpenClaw, a unified agent framework supporting daily assistance, office productivity, and deep research workflows. We evaluate Nanbeige4.2-3B across three representative local-assistant scenarios:

\begin{itemize}[leftmargin=10pt]
    \item \textbf{Daily Tasks.}
    PinchBench-V2 and ClawGym assess common personal-assistant capabilities including task execution, information retrieval, tool interaction, and everyday workflow automation.
    
    \item \textbf{Office Tasks.} 
    GDPval and AgentIF-Oneday evaluate document understanding, office productivity, and multi-step workplace workflows.
    
    \item \textbf{Deep Research.}
    DeepResearch Bench II~\cite{li2026deepresearchbenchii} and ResearchRubrics~\cite{sharma2025researchrubrics} cover long-horizon planning, iterative information retrieval, evidence synthesis, and research-oriented reasoning across multiple external resources.
\end{itemize}

\begin{table}[t]
\centering
\small
\setlength{\tabcolsep}{5.0pt}
\renewcommand{\arraystretch}{1.08}
\caption{Evaluation results on local personal assistant benchmarks under the OpenClaw framework.}
\label{tab:local_assistant}
\begin{tabular}{lccc}
\toprule
Benchmark
& \makecell[c]{Qwen3.5-4B}
& \makecell[c]{Qwen3.5-9B}
& \makecell[c]{Nanbeige4.2-3B} \\
\midrule

\multicolumn{4}{c}{\textit{Daily Tasks}}\\
\midrule

Pinch-Bench-V2          & 63.9 & \underline{68.2} & \textbf{74.7}\\
Claw-Gym                & 53.0 & \underline{56.1} & \textbf{65.0}\\

\midrule

\multicolumn{4}{c}{\textit{Office Tasks}}\\
\midrule

GDPval                  & 37.0 & \underline{38.0} & \textbf{68.8}\\
AgentIF-Oneday         & 27.0 & \underline{32.1} & \textbf{58.9}\\

\midrule

\multicolumn{4}{c}{\textit{Deep Research}}\\
\midrule

DeepResearch Bench II   & 26.0 & \underline{28.3} & \textbf{33.4}\\
ResearchRubrics         & 35.1 & \underline{37.2} & \textbf{44.8}\\

\bottomrule
\end{tabular}
\label{tab:sub_benchmarks}
\end{table}

\paragraph{Overall Results.}

Table~\ref{tab:sub_benchmarks} reports results obtained with the same OpenClaw scaffold, tool set, and protocol for all models. Nanbeige4.2-3B outperforms Qwen3.5-4B and the larger Qwen3.5-9B on all six benchmarks spanning daily tasks, office workflows, and deep research. The gains are consistent across all three task categories and are especially clear on office workflows, indicating that the model's general agentic capabilities transfer well to a unified local-assistant setting. These results support its use as a compact local personal assistant.

%% file: sec_conclusion.tex
\section{Conclusion}

We present Nanbeige4.2-3B, a compact model that combines broad agentic capabilities with strong reasoning under a 3B non-embedding parameter budget. Through a Looped Transformer architecture, pre-training from scratch on an improved 28T-token corpus, and scalable SFT trajectory construction and multi-stage RL recipes, Nanbeige4.2-3B achieves solid performance across code-agent, office-agent, and complex tool-use tasks while remaining highly competitive in mathematical, coding, and scientific reasoning. Its performance in local personal-assistant workflows further demonstrates the practical potential of compact agentic models.

Looking forward, we will continue improving model capacity and efficiency under a constrained computation budget. On the architecture side, building on Nanbeige4.2, we are improving how layers are organized within the loop, how information is passed and reused across different depths, and how token embeddings incorporate local n-gram context as external memory.\footnote{Preliminary implementations are already included in the modeling code released with Nanbeige4.2.} These directions aim to improve model performance with minimal additional FLOPs. We are also exploring linear and sparse attention mechanisms for more efficient long-context modeling. On the agent side, we will continue scaling post-training along three dimensions: task difficulty, the number and diversity of environments, and trajectory quality. Other than post-training, we plan to study how agentic capabilities can be established earlier during pre-training. We will train and release larger Nanbeige models and investigate how to use them to distill a stronger small model.

%% file: sec_appendix.tex
\section{Author List}

Authors are listed in \textbf{alphabetical order by first name}. Names marked with an asterisk (*) denote individuals who were previously affiliated with our team. Yang Song is the corresponding author and can be reached at \texttt{songyang@kanzhun.com}.

Chen Yang, Chengrui Huang, Fufeng Lan, Hanhui Chen, Hao Zhou, Huatong Song\textsuperscript{*}, Jiaqi Cao\textsuperscript{*}, Jiaying Zhu, Jinlin Niu, Kai Wang, Lisheng Huang, Qiliang Liang, Ran Le, Ruixiang Feng\textsuperscript{*}, Shuang Sun\textsuperscript{*}, Tao Gu, Tao Zhang, Tianyu Luo, Yang Song\textsuperscript{\dag}, Yun Xing, Yuntao Wen\textsuperscript{*}, Ziyao Xu, Zongchao Chen, Zongqiang Li

\section{Evaluation settings}
\label{app:evaluation}

\subsection{General Inference Settings.}

Unless otherwise specified, we evaluate Nanbeige4.2-3B with the following configuration:

\begin{itemize}[leftmargin=1.5em, itemsep=0.2em, topsep=0.4em]
    \item Temperature: 0.6
    \item Top-p: 0.95
    \item Top-k: 20
    \item Context Window: 256k tokens
\end{itemize}

\subsection{Code Agent Evaluation}
\emph{SWE-bench Verified}: We evaluate models using the OpenHands framework as the agent scaffold with a context window of 256k tokens and a maximum output limit of 32k tokens. The decoding temperature is set to 1.0. Each task run is allocated a 4-hour timeout and capped at a maximum of 250 interaction turns. Final performance metrics are averaged over 8 independent runs.

\emph{SWE-bench Pro}: We evaluate models using the SWE-agent harness with a context window of 256k tokens and a maximum output limit of 32k tokens. The decoding temperature is set to 1.0. Given the increased complexity of long-horizon tasks, each run is granted a 10-hour timeout and a maximum limit of 250 interaction turns. Final results are averaged over 8 independent runs.

\emph{Terminal-Bench 2.0}: We evaluate Terminal-Bench 2.0 using the Harbor/Terminus-2 framework with a JSON output parser. Models are evaluated with a 256k context window, a maximum output limit of 32k tokens, and a temperature of 1.0. Each run is allocated 8 CPU cores, 24GB of RAM, a 4-hour timeout, and a maximum of 250 interaction turns. All reported results are averaged over 8 independent runs.

\subsection{General Agent Evaluation}

\emph{GDPval Rubrics}: We evaluate models using our in-house harness, which provides web-search tools and a sandboxed environment. Final outputs are scored by an agent acting as a judge in the same environment. The judge evaluates each rubric and normalizes the aggregate score to 0--100.

\emph{AgentIF-Oneday}: We follow the same evaluation protocol as for GDPval Rubrics.

\emph{OfficeQA-Pro}: We evaluate models in the most challenging setting using our in-house harness: each question is paired with the complete collection of original PDF materials, without any parser and indication of which documents are relevant.

\emph{Claw-Eval}: 
We evaluate and report results only on the 157 general tasks.
We preserve the complete reasoning context during model evaluation~(e.g., The reasoning-parser of the sglang service is not turned on to ensure that the thought process remains in context). For final scoring, content preceding the closing \texttt{</think>} tag is removed. We use DeepSeek-V4-Pro as the judge model and impose an overall task timeout of 3,600 seconds.

\emph{MCP-Atlas}: During evaluation, we enable the system prompt and remove content preceding the closing \texttt{</think>} tag for final scoring. We use GLM-5.1 as the judge model, set the per-call and overall task timeouts to 1,200 and 3,600 seconds, and permit at most 20 tool-use rounds per task.

\subsection{OpenClaw-Based Evaluation}

To maximize the model's potential, we retain the complete reasoning content throughout evaluation. To enable this setting, the model need to be added to OpenClaw's reasoning-content replay whitelist.

\emph{PinchBench-V2}: We use OpenClaw's default free DuckDuckGo Search backend. Each task is allotted a timeout of 10,800 seconds. We use Qwen3.7-Plus as the judge model.

\emph{Claw-Gym}: We adopt the same evaluation setup as for PinchBench-V2.

\emph{GDPval Rubrics} and \emph{AgentIF-Oneday}: We adopt the same generation setup as for PinchBench-V2. We use Qwen3.7-plus for rubric scoring on the products of the task.

\emph{DeepResearch Bench II}: We use Brave Search as the search backend. Each task is allotted a timeout of 3,600 seconds and a maximum of 200 agent turns. We use DeepSeek-V4-Flash as the judge model with \texttt{temperature}$=0$, and follow the official judge prompt and score-aggregation procedure.

\emph{ResearchRubrics}: We adopt the same evaluation setup as for DeepResearch Bench II, while following the official ResearchRubrics judge prompt and benchmark-specific evaluation protocol.

\section{Acknowledgments}

We thank Guoliang Cheng, Jianyong Xia, Kewen Zhu, Shu Xu, Yanxi Xie, and Zhihao Li for their discussions and involvement in this project.